# Dynamic Cognition Applied to Value Learning in Artificial Intelligence

# Cognição Dinâmica Aplicada à Aprendizagem de Valor em Inteligência Artificial

Nicholas Kluge Corrêa, PhD. Candidate
PUCRS[1]

Nythamar Fernandes De Oliveira, PhD. Professor
PUCRS, CNPq Fellow[2]

185

## ABSTRACT

Experts in Artificial Intelligence (AI) development predict that advances in the development of intelligent systems and agents will reshape vital areas in our society. Nevertheless, if such an advance isn't done with prudence, it can result in negative outcomes for humanity. For this reason, several researchers in the area are trying to develop a robust, beneficial, and safe concept of artificial intelligence. Currently, several of the open problems in the field of AI research arise from the difficulty of avoiding unwanted behaviors of intelligent agents, and at the same time specifying what we want such systems to do. It is of utmost importance that artificial intelligent agents have their values aligned with human values, given the fact that we cannot expect an AI to develop our moral preferences simply because of its intelligence, as discussed in the Orthogonality Thesis. Perhaps this difficulty comes from the way we are addressing the problem of expressing objectives, values, and ends, using representational cognitive methods. A solution to this problem would be the dynamic cognitive approach proposed by Dreyfus, whose phenomenological philosophy defends that the human experience of being-in-the-world cannot be represented by the symbolic or connectionist cognitive methods. A possible approach to this problem would be to use theoretical models such as SED (*situated embodied dynamics*) to address the values learning problem in AI.

[1] E-mail: nicholas.correa@acad.pucrs.br, Orcid: https://orcid.org/0000-0002-5633-6094
[2] E-mail: nythamar@yahoo.com, Orcid: https://orcid.org/0000-0001-9241-1031

## RESUMO
Especialistas em desenvolvimento de Inteligência Artificial (IA) preveem que os avanços no desenvolvimento de sistemas e agentes inteligentes remodelarão áreas vitais em nossa sociedade. No entanto, se tal avanço não for feito com prudência, ele pode resultar em resultados negativos para a humanidade. Por esta razão, vários pesquisadores da área estão tentando desenvolver um conceito robusto, benéfico e seguro de inteligência artificial. Atualmente, vários dos problemas abertos no campo da pesquisa da IA surgem da dificuldade de evitar comportamentos indesejados de agentes inteligentes, e, ao mesmo tempo, especificar o que queremos que tais sistemas façam. É da maior importância que agentes inteligentes artificiais tenham seus valores alinhados com os valores humanos, dado que não podemos esperar que uma IA desenvolva nossas preferências morais simplesmente por causa de sua inteligência, como discutido pela Tese de Ortogonalidade. Talvez esta dificuldade venha da maneira como estamos abordando o problema de especificação de objetivos, valores e fins, ao utilizar métodos cognitivos representacionais. Uma solução para este problema seria a abordagem cognitiva dinâmica proposta por Dreyfus, cuja filosofia fenomenológica defende que a experiência humana de estar-no-mundo não pode ser representada por métodos cognitivos estritamente simbólicos ou conexionistas. Uma abordagem possível para este problema seria a utilização de modelos teóricos como o SED (*situated embodied dynamics*) para abordar o problema de aprendizagem de valores em IA.

## PALAVRAS-CHAVE
Inteligência artificial; Aprendizagem de valores; Ciência cognitiva; Cognição dinâmica

## INTRODUCTION

Researchers in Artificial Intelligence (AI) development stipulate that within ten years many human activities will be surpassed by machines in terms of efficiency. Several aspects of our public policies will need to be modified to accommodate such advances, which promise to reshape areas such as transportation, health, economics, military fighting, lifestyle, etc. There is also concern about the risks that machines with a high level of human or superhuman intelligence may bring to humanity in the coming decades. A survey conducted by Müller and Bostrom (2016) consisted of building a questionnaire to assess progress in the field of AI research and prospects for the future, interviewing various experts in the field. The questionnaire showed that, on average, there is a 50% chance that high-level (human) machine intelligence will be achieved between 2040 and 2050, reaching a 90% probability by 2075. It is also estimated that this intelligence will exceed human performance in two years (10% chance) to 30 years (75% chance) after reaching human levels of intelligence.

However, in the same survey, 33% of respondents classified this development in AI as "bad" or "extremely bad" for humanity. As there is no guarantee that such systems will be "good" for mankind, we should investigate further the future of artificial intelligence and the risks it poses to the human race. Some several open questions and problems need to be solved. How will we remedy the economic impacts of AI to avoid negative effects such as mass unemployment (FREY; OSBORNE, 2013)? How can we prevent the automation of jobs from pushing the distribution of income into a law of disproportionate power among classes, genders, and races (BRYNJOLFSSON; MCAFEE, 2014)? Can autonomous lethal weapons be built without changing humanitarian rights, and should autonomous weapons be completely banned (DOCHERTY, 2012)? How can we ensure privacy by applying machine learning to confidential data such as medical data sources, phone lines, emails, and online behavior patterns (ABADI *et al.*, 2016)?

Some researchers have already created models (ASI-PATH) of how an AI could pose an existential threat, becoming super-intelligent through recursive self-improvement (BARRET; BAUM, 2017), something known in the AI literature as a Singularity. Such models suggest scenarios where intelligent agents after obtaining some kind of strategic advantage, such as advances in nanotechnology or robotics, could achieve considerable power of domination over our environment (BOSTROM; ĆIRKOVIĆ, 2008). Shulman (2010, p. 2) suggests a model that explains in which situations an AI would abandon cooperation with the human society, and take a hostile action. In it, an artificial agent that believes it has a P probability of being successful, if it initiates aggression, receiving some expected utility [EU (Success)], and with a (1 - P) probability of failing, receiving [EU (Failure)]. Contrary, if it gives up the aggressive strategy the agent will receive utility [EU (Cooperation)]. The AI will rationally initiate the aggression if, and only if:

$$P \times EU(success) + (1 - P) \times EU(failure) > EU(cooperation)$$

The development of an AI ethic presupposes the intuitive formulations of Isaac Asimov's so-called Three Laws of Robotics (1950), at a time when this theme still seemed relegated to the realm of science fiction. Recalling that such ethical-moral codifications were introduced in a 1942 tale, Runaround: (1) A robot may not harm a human being or, by inaction, allow a human being to be harmed; (2) a robot must obey the orders given by human beings, except where such orders conflict with the First Law; (3) a robot must protect its existence, provided such protection does not conflict with the First or Second Law. Currently, this ethical orientation of "no harm to mankind" is extended not only to robots and robotic artifacts but to machines and intelligent devices generally associated with AI resources. But would simple deontological laws like the ones cited above be enough to ensure safe behavior?

187

## 1 SAFETY ISSUES IN AI

Ultimately, there is a consensus in the literature: AI development must be done in a safe, beneficial, and robust manner. An article published by Amodei et al (2016), entitled "Concrete Problems in AI Safety", and lists several open problems in the field of AI research that must be addressed if we are to reap the benefits of AI without compromising our safety. These problems are classified into specification and robustness problems, which are the current barriers to be overcome in the area (LEIKE *et al.,* 2017). To better synthesize and develop the content of this study, we will refer briefly only to specification errors. Specification errors occur when the utility function of the AI is poorly specified by programmers, causing unwanted and even harmful results, even if the learning is perfect with explicitly clear data. Some examples of specification errors are:

- *Negative side effects*: those occur when the maximization of the reward function focuses on achieving a goal while the agent ignores important factors in the environment, causing potential side effects;
- *Reward Hacking*: the AI agent finds a solution to its goal that maximizes its reward function, but unexpectedly, perverting the intention of the programmers;
- *Corrigibility*: this problem concerns how we can be able to interrupt an agent if it is behaving unexpectedly since expected utility maximizers have instrumental convergent goals about utility preservation that we still don't know how to "shut-down".

Two theses published by Bostrom, (2012), firstly proposed by Omohundro (2008) in his seminal paper "The Basic AI Drives", point out how these problems can present a risk. The Thesis of Instrumental Convergence shows us how a series of self-improvement and self-preservation goals can be pursued by almost any intelligent agent with a terminal goal. We can formulate this thesis as follows:

> Several instrumental objectives can be identified, which are convergent in the sense that their attainment would increase the chances of the agent's terminal objective, implying that these instrumental objectives are likely to be pursued by any intelligent agent (BOSTROM, 2012, p. 6).

Without careful engineering of these systems, risks with an "intelligence explosion" (the exponential increase in the cognitive capacity of the agent) can create agents much more powerful than our ability to control them. On the other hand, and correlated to the first thesis, the Orthogonality Thesis proposes that intelligence and terminal goals have independent and orthogonal properties:

> Intelligence and ultimate goals are orthogonal axes along which possible agents can freely vary. In other words, more or less any level of intelligence could, in principle, be combined with more or less any final objective (BOSTROM, 2012, p. 3).

The thought behind the orthogonality thesis is analogous to the so-called Hume's Guillotine (also known in English as Hume's fork or Hume's law), opposing what is factually and empirically verifiable (matters of fact and real existence) to what should be, in rational terms, normative and counterfactual (relations of ideas). Hume observed a significant difference between descriptive statements and prescriptive or normative statements, and therefore, it would not be obvious, self-evident (self-evident) or valid (valid) to derive the latter from the former. The undue passage from being (Is) to being (Ought), which would be one of the seminal problems of research in metaethics, normative ethics, and applied ethics in the twentieth century, was noted by the Scottish philosopher in a famous passage in section I of part I of his Treatise of Human Nature:

> In every moral system I have encountered to date, I have always noticed that the author follows for some time the common way of reasoning, establishing the existence of God, or making observations regarding human affairs, when suddenly I am surprised to see that, instead of the usual propositional copulations, as it is and is not, I do not find a single proposition that is not connected to another by one should or should not. This change is imperceptible but of the utmost importance. For as this must or must not express a new relationship or affirmation, it would need to be noted and explained; at the same time, it would need to give a reason for something that seems inconceivable, that is, how this new relationship can be deduced from entirely different ones (HUME, 2009, p. 509).

Just as descriptive and purely factual statements can only bind or imply other descriptive or factual statements, and never norms, the problems of orthogonality and value alignment consist in guaranteeing that an advanced AI, if it develops enough intelligence to gain power over the human species, that such intelligence would do with human beings only what we would wish or accept it to be done.

In this sense, the problem of alignment is identical to what we see in moral philosophy about utilitarianism, in that the maximization of utility by some moral agent can culminate in morally repugnant conclusions, including the violation of the rights of others. Although it may guarantee the resolution of tasks in computational time (polynomial), the mere efficiency or optimization of procedures does not ensure normative universalizability (as it would be, moreover, a basic premise of ethical deontological and non-utilitarian models) and may eventually conflict with the interests or rights of other people. We should also note that the ethics of artificial intelligence is part of the ethics of technology in general and, specifically, for robots, learning machines, and other artifacts and artificially intelligent entities.

AI ethics comprises both robotics (robotic ethics), which is concerned with the moral behavior of human beings when designing, building, using, and programming artificially intelligent beings, and a machine ethic, which is concerned with the moral behavior of artificial moral agents themselves. Ethics, in general, have much to learn, to teach, and to interact with the ethics of artificial intelligence, especially through the ethical-normative challenges of orthogonality, value alignment, and transhumanism, integrating the neurobiological, cultural, and technological legacies of the *homo sapiens sapiens.*

Practically all problems of specification, robustness, and goal alignment and value specification seem to occur at the same point, that is, when our representations of values and goals lose their meaning or are misinterpreted. Is the objective-representational approach doomed to error? Would the cognitive models used in the creation of artificially intelligent agents, especially symbolism and connectionism, be incapable of expressing the meaning of human values? If so, would there be any alternative?

## 2 COGNITIVE MODELS: SYMBOLISM AND CONNECTIONISM

Since the late 50s, the discussion about cognition and intelligence has been permeated by the computational framework, also known as the symbolic view. This perspective starts from the assumption that cognitive systems are intelligent in that they can encode knowledge into symbolic representations. Symbolists believe that through sets of "if-then" rules and other forms of symbolic manipulation, all forms of cognition can be accomplished (THAGARD, 1992). Allen Newell defined the symbolic view, which is also referred to as the Physical Symbol System Hypothesis, as follows:

> Natural cognitive systems are intelligent by being physical systems that manipulate symbols in such a way as to present intelligent behavior, codifying knowledge about the external world in symbolic structures (NEWELL, 1990, p. 75).

Newell has dedicated much of his work to building systems that express his vision of a physical symbol system. His most promising model is known as SOAR. SOAR is a symbolic computational system that formulates its tasks based on symbol and goal hierarchies, thus generating a decision making system for problem-solving.

But for connectionists, the phenomenon of cognition is a high-level effect that depends on lower-level phenomena. Thus, the connectionist hypothesis encapsulates the idea that the most fundamental characteristic of a cognitive agent is not its capacity for symbol manipulation, but its architecture. Thus, connectionists attack the problem of cognition by reverse-engineering the human central nervous system, and copying its basic processing unit, the neuron (CHURCHLAND; SEJNOWSKI, 1992, p. 2). Sejnowski (1988, p. 7) notes in his connectionist hypothesis: "The intuitive processor is a dynamic sub-conceptual connectionist system that does not admit a complete, formal and precise description on a conceptual level".

The theories of cognition just cited (symbolism and connectionism) can be considered theoretical structures that provide us with the filters, analogies, and metaphors by which we try to understand the phenomenon of cognition, and thus create theoretical models that can generate simulations to be tested. Symbolism highlights the internal representations of the system and the algorithms by which these representations are manipulated. Connectionism emphasizes the neural network architecture and the methods (algorithms) of learning.

However, the limitations of the symbolic computational hypothesis, especially in the aspects of time, architecture, computing, and representation, led researchers to consider new theoretical models, such as the dynamic hypothesis of van Gelder (1998). In this article, we will explore some of the consequences of adopting a dynamic cognitive approach to the problem of AI and value alignment. Even so, we would like to make clear that the authors don't endorse an anti-representational position. Humans constantly use and manipulate representations, as in language, writing, speech, music, and other forms of abstract thinking. However, we skeptically position ourselves concerning the function of representations in systems that involve values and objectives, and therefore, goal-oriented behavior.

## 3 A CRITIQUE OF THE SYMBOLIC METHOD

One of the biggest criticisms raised against the symbolic model for cognition is the difficulty in meeting time constraints. When trying to replicate the phenomenon of cognition van Gelder and Port (1998, p. 2) states that the symbolists *leave time out of the picture*. Since the objective of cognitive science is to describe the behaviour of natural cognitive agents, agents that operate in real-time, a cognitive model that replicates the human experience of cognition must present real-time cognitive processes.

The limits imposed by symbolic architecture are another reason for criticism of the symbolic method. For Newell (1990, p. 82), the behavior of a cognitive system is determined by the variables being processed by a fixed structure, which is its architecture. Dynamists criticize this view of the cognitive system as "*a box*" within a body, in turn within a physical environment. However, where do we draw the line that divides the box from your body? And more controversially, the body with the environment? Van Gelder and Port, (1998, p. 8), analyze the internal architecture in the cognitive agent as not being a fixed structure, where all aspects of cognition,

191

brain-body-environment, mutually influence each other continuously and dynamically.

Consequently, this view of fixed architecture often makes people refer to the symbolic approach to cognition as the computational method since it describes the mind as a type of computer. In this characterization, the body, through the sensory organs, delivers to the cognitive system (brain) representations of the state of its environment; the system on its part calculates an appropriate response, and the body carries the action (VAN GELDER, PORT, 1998, p. 1). However, this system of perceiving-planning-acting ignores phenomena in decision makings, such as reflex actions, and the speed with which such actions are expressed in real cognitive agents, showing once again that the symbolic computational method has no basis with the biological and physical reality of the phenomenon of cognition.

Hubert Dreyfus (1992) was one of the most prominent critics of the symbolic representational approach in the field of AI research. Based on the hermeneutic-existentialist philosophy proposed by Martin Heidegger, Dreyfus indicated in his works that the manipulation of symbols and representations is not enough to generate the non-representational type of existence of a being in the world (Dasein). At the bottom of this impasse, there remains a criticism of materialist Cartesian thought and subject-object dualism: materialist Cartesianism that attempts, without success, to replicate the whole world "inside the mind" is doomed to fail according to Dreyfus, because it is impossible to contain the world inside the mind for the simple fact that the world is infinitely complex and we are finite creatures (DREYFUS, 2007). Thus, a self-contained, rigid system is not capable of duplicating the type of cognitive agent we desire. Perhaps this indicates to us that representations and experience must operate together for the former to have meaning.

## 4 CONNECTIONISM AND VALUE LEARNING

We can see that many of the problems related to misspecification come from the difficulty of programmers in expressing the meaning of what is proposed by the language (specification errors) and how this should change when the context of the environment evolves (robustness errors). Be it the representative cognitive model, using rules of behavior, or the connectionist model, using artificial neural networks with reward functions, we still reach the same impasse. How to express our goals and align the values of artificially intelligent agents with ours?

The connectionist approach to cognition encounters several difficulties in this task. Commonly artificial neural networks are trained in a supervised manner using labeled training data. However, this method may not be the safest for value learning. Dreyfus and Dreyfus (1992) cite an example, one of the most famous anecdotes of machine learning literature, where a machine learning system was trained to classify military ground vehicles hidden among the trees. The classifier during the training was able to identify with great precision the desired vehicles. However, the system had a poor performance with images outside the training group during its deployment phase. It was later discovered that the set of photos containing vehicles used in training were all taken on a sunny day, while the images without the vehicles were made on a cloudy day. What the classifier was identifying was the brightness of the image and not the presence of military vehicles. Potentially, learning values by supervised learning may be susceptible to this failure mode.

Besides supervised training methods, reinforcement learning techniques, which use utility functions as a proxy for desirable results, are extremely poor in identifying ambiguities (SOARES, 2016). We also call this type of problem Sorcerer's Apprentice, which are situations where the system, due to divergence in testing environments and new environments, and also goal misspecification, has the opportunity to hack its reward or optimize it perversely. The reward hacking scenario can be exemplified as follows: imagine a cleaning robot whose reward proxy is how much dirt it sucks up and fills its container. If its reward function was just that, maximizing how much dirt is fed to the container, the agent could adopt a policy of filling, emptying, and refilling with the same dirt, in an infinite cycle, its container, and not cleaning the AI developer's office.

An alternative would be to model the intent of operators using inverse reinforcement learning (NG; RUSSELL, 2000): where one agent tries to identify and maximize the reward function of some other agent in the environment (usually a human operator). The concept of the utility function is a mathematical formalization for the notion of human values, or a normative rule, and is widely used in economics and decision theory. However, one of the best-known problems of this model is the empirical fact that humans violate many axioms of utility theory and do not have consistent utility functions (TVERSKY; KAHNEMAN, 1981). In this way, optimally inverse reinforcement learning demonstrates the problem of learning "errors" or biases in human behavior as valid solutions.

Moreover, situations where humans are part of the reward system of an AI, also called human-in-the-loop, are not considered safe. There would be strong incentives for the agent to manipulate the human part of its reward mechanism if it meant an increase in reward (HIBBARD, 2012; BOSTROM, 2014). In general, our current training methods for the connectionist cognitive model are not appropriate for an AI or AGI operating in the real world. Possible scenarios of self-improvement as explained by the Instrumental Convergence thesis can generate undesired consequences. The ultimate goal of these agents is to maximize the reward, being our values and goals only instrumental to this ultimate goal. Such agents can learn that human goals are instrumentally useful for high rewards, but replaceable, especially if the intelligence of these agents is superior to ours.

Whether by symbolic or connectionist models, so far human goals cannot be safely and robustly expressed. Given the importance of value alignment with AI new methods must be investigated. We propose in this article that the dynamic cognitive model offers a new way of thinking about the problem of alignment. In the following section, we will discuss some of the characteristics of the theoretical dynamic model for cognition.

## 5 DYNAMIC COGNITION

It can be stated that many theoretical models begin as metaphors or analogies, later becoming theories that can be implemented in models and subsequently simulated. The conceptual structures that we form through this process can have a great impact on the way we conduct our studies, the way we approach the problem, the language we describe the phenomena, and the way we formulate a question and interpret the answer. The theory of dynamic systems invites us to think about the phenomenon of cognition and human experience in a progressive way, as proposed by van Gelder (1998, p. 4): "Natural cognitive systems are certain types of dynamic systems, and are best understood from the dynamic perspective". Dynamic systems, in this sense, are systems in which, as they evolve in time, their variables are continuously and simultaneously determining the evolution of each other, in other words, they are systems governed by nonlinear differential equations (VAN GELDER; PORT, 1998, p. 6). With this statement, the dynamist puts the agent in a situation of coupling with the environment, turning the brain-body-environment into an autonomous cognitive dynamic system. In this view, it no longer makes sense to talk about cognition or experience without recognizing the three aspects of this triad (VAN GELDER; PORT, 1998, p. 23).

The situated activity has its philosophical origins in the phenomenological work of Heidegger (2012), which Dreyfus (1992) applied to the field of AI. Dreyfus proposed that the Heideggerian agent couldn't be separated from its environment or its interpretative context. Gibson's Ecological Psychology (1979) is also a precursor of situated activity, with its notion of affordances: Gibson emphasizes the environment-organism relationship in the phenomenon of perception as a two-way street, where one perceives to act, and acts to perceive. The idea of situated cognition can be extended to theories such as *extended cognition* (CLARCK; CHALMERS, 1998; ROCKWELL, 2010), which invites us to think differently, opposing the Cartesian thought that places the mind imprisoned inside the brain. We explain gravity as the relationship between gravitational fields; electromagnetism by electromagnetic fields; the position of subatomic particles is expressed through probabilistic waves using Schrödinger's equation, De Broglie's wavelength, and Heisenberg's uncertainty principle. Thus, it seems likely that a sophisticated theory for explaining cognition, and consciousness, and experience should involve the dynamic fluctuation of fields.

The theoretical model we present in this article is SED (situated embodied dynamics), proposed by Beer (2000), which emphasizes how the cognitive experience arises from the dynamic interaction brain-body-environment. In the first place, SED takes into account the situation as being fundamental to cognition, placing concrete action, which is, literally acting in the world, as something more fundamental than the abstract descriptions of this action. Thus, the final work of the intelligent agent is to act, an action that occurs in an environment, which is a central part of the phenomenon since it is what gives meaning and context to the action. And the interaction of the agent with the environment is mutual, not being the environment just a source of problems to be solved, but a partner with whom the agent is involved from moment to moment. In the SED approach, the concept of embodiment says that the physical form and its functional and biomechanical aspects are essential aspects for behavior, as well as its biology and physiology, in the case of artificial agents, mechanics, hardware, and software. All these factors create the conceptual realization by which the agent creates its experiences and representations.

The thought of embodiment has its origin in the phenomenological work of Merleau-Ponty (1962), who was moreover one of the forerunners of Gibson's notion of affordance, placing body involvement as crucial to the way we perceive and act with the environment. Also, being the biological structure that supports the cognitive phenomenon, we must think about the implications or possibilities of this phenomenon being generated by electronic components, given the importance of the embodied experience in the creation of abstract concepts (LAKOFF; JOHNSON, 1999). Thus, the role of language, metaphors, and mental representations in the formulation of concepts used in scientific theories is evident, despite all ontological commitment to a certain scientific realism. The term "naturalized epistemology", forged by W.V. Quine in his 1969 seminal essay "Epistemology Naturalized", followed several of the epistemic premises of Hume's skepticism, which, as we pointed out above, solves every platonically inspired foundation, including the dualism of Cartesian rationalism, in its pretension to justify a sure knowledge of the truth of the outside world. According to Quine (1969):

195

> It was sad for epistemologists, Hume and others, to have to agree on the impossibility of strictly deriving the science of the external world from sensory evidence. Two fundamental principles of empiricism remained unassailable, however, and remain so today. One is that any evidence that exists for science is sensory evidence. The other is that any inculcation of word meanings must ultimately rest on sensory evidence (p. 75).

As in Quine, the Humean-inspired empiricism that interests us, from Dreyfus, Rorty, Prinz, and neopragmatism, is intersubjective, falsificationist, and interestingly. externalist. That is, a form of social linguistic and historically co-constitutive pragmatism of the observer subject and the objective world to be known, experienced, and lived. The problem of knowledge, as well as that of giving reasons for moral action, remains the great human problem. According to the Humean formulation, in the words of Quine (1969, p. 72), "the Humean problem is the human predicament" so that not even induction (such as that which has been adopted by models of reflexive balance in metaethics and philosophy of science) can solve the naturalistic fallacies that arise from the guillotine. The externalism of the naturalists, in the wake of Hume and Quine, would here oppose the internalism of the rationalists and Kant, according to which the epistemic justification for cognition and moral action is found in consciousness (*cogito*) or a structure of transcendental subjectivity.

Although we cannot develop here the internalist-externalist problem, we believe that the debate between rationalism and empiricism that preceded it authorizes us to assert, as Quine suggested, that Hume's great mistake would have been to reduce analytical judgments to a priori, universal, necessary judgments, as opposed to synthetic ones. In turn, they are reducible to posterior judgments, contingent particularities. Without solving the problem of induction, but on the contrary, allowing their return through the back door, as Popper would show, by the self-deception of those who intend to justify the moral action with a transcendental or normativism argument. Our programmatic intuition on AI ethics is, therefore, that (i) neither naturalism seems to be able to reduce the alignment to a utilitarian program, nor (ii) the deontological normative models nor their transcendental arguments seem satisfactory to avoid anthropomorphic suspicion.

Work in the field of autonomous robotics emphasizes that intelligent behavior is an emerging property of an agent incorporated in an environment with which it must interact continuously (CHIEL; BEER, 1997). Thus, the symbolic view of cognition, which places the brain as the source of commands that are issued to the body, may be incomplete. There may be a cognition or "mind" of the body (or mechanical system), governed by the laws of physics itself. This vision puts the nervous system not in a position to issue commands, but suggestions, reconciled with the biomechanical and ecological context. There is the possibility that an AI that has an understanding of human concepts would require a design very close to that of a human being, an anthropomorphic design.

Finally, to understand the SED approach we must analyze the assumed dynamics. We refer to dynamics as a mathematical theory that describes systems that systematically change over time. The dynamic framework also provides us with a different filter to observe the phenomenon in question. The most common examples of dynamic systems are sets of partial differential equations, used to describe phenomena such as the movement of water, behavior of electromagnetic fields, the position of subatomic particles among other natural phenomena. Thus, the dynamic perspective brings with it a set of concepts and filters that influence the way we think about the phenomenon studied. When approaching any system from the dynamic perspective, we try to identify a set of state variables whose evolution can explain the observed behavior, the dynamic laws by which the values of these variables evolve in time, the dimensional structure of their evolution, possible states and dominant attractors (BEER, 1998; 2000; 2003). Thus, we believe that the dynamic approach is more suitable to represent the phenomenon of normative and cognitive agency.

## DISCUSSION AND CONCLUSION

How can this dynamic approach to cognition be useful for the problem of value alignment? As we have seen in this study, the imminent advance of AI technologies, and the importance that such progress is made prudently, comes from the fact that we can't anthropomorphize AI, and expect artificial intelligent agents to have the same terminal objectives (values) as us. Therefore, value learning becomes an area of crucial importance in the field. The limitations present in the symbolic method, and the connectionist model, maybe indicating that a different approach to the problem of cognition and normativity should be considered. Dynamism approaches this problem differently and unveils new aspects that both the symbolic and connectionist models leave aside.

How should we understand the nature and role of this inner state within a dynamic agent? The traditional computational interpretation of such states would be as internal representations. Unfortunately, despite the fundamental role that the notion of representation plays in computational approaches, there is very little agreement about what its real function is in controlling and maintaining behavior. We should also remember that symbolism, connectionism, and dynamism are theoretical structures, not scientific theories of the natural world. That is, they cannot be proved or refuted. While symbolism emphasizes the manipulation of internal representations, connectionism emphasizes the architecture of the network and the training protocol. The dynamic view, on the other hand, highlights the trajectory space and the determining influences on the brain-body-environment system. However, as stated above, we do not put ourselves in a position of anti-representationalism. On the contrary, a complete theory of cognition is likely to use all three theoretical structures. We suggest that in certain cases the internal functioning of a cognitive and normative agent cannot be interpreted as symbolically representative unless we redefine what a representation really can be or mean.

The dynamic approach differs from the symbolic and connectionist cognitive models because it places biomechanics and ecology with the same relevance as neural activity in the phenomenon of cognition. Perhaps the difficulties we have encountered in learning values and other problems in the field of AI are because we are ignoring two crucial factors of the phenomenon. The implications of the dynamic hypothesis not only bring a new way of thinking but also new problems to the field of AI research, thus nurturing new ideas in areas such as neurophilosophy, neuroscience, metaethics, and cognitive science. In conclusion, improvement and a better understanding of dynamic systems concepts are needed, with the promise that such methods can be useful for the problem of value alignment in AI and the cognitive science community in general.

199